\documentclass{article}
\usepackage[british]{babel}
\usepackage[utf8]{inputenc}
\usepackage{rotating}
\usepackage{amsthm}
\usepackage{amssymb}
\usepackage{amsmath}
\usepackage{nicefrac}
\usepackage{float}
\usepackage{setspace}
\usepackage{url}
\usepackage{textcomp}
\usepackage{bm}
\usepackage{mathtools}
\usepackage{makecell}
\usepackage{caption}
\usepackage{subcaption}
\usepackage{enumerate}
\usepackage{listings}
\usepackage{tabu}
\usepackage{color}
\usepackage{cprotect}
\usepackage{graphicx}
\usepackage{makeidx}
\usepackage{float}
\usepackage[ruled]{algorithm2e}
\usepackage[toc,page,title]{appendix}
\usepackage{afterpage}

\title{Methods for segmenting cracks in 3d images of concrete: A comparison based on semi-synthetic images}
\author{Tin Barisin$^{a,b}$ \and Christian Jung$^{b}$ \and Franziska Müsebeck$^b$ \and Claudia Redenbach$^b$ \and Katja Schladitz$^a$}

\date{
	{\small $^a$Fraunhofer Institut für Techno- und Wirtschaftsmathematik \\ Fraunhofer-Platz 1, 67663 Kaiserslautern, Germany}\\
	\vspace{0.5em}
	
	{\small $^b$Technische Universität Kaiserslautern \\  Gottlieb-Daimler-Straße 48, 67663 Kaiserslautern, Germany}
}

\begin{document}

\maketitle

\textbf{Abstract}\\
Concrete is the standard construction material for buildings, bridges, and roads. As safety plays a central role in the design, monitoring, and maintenance of such constructions, it is important to understand the cracking behavior of concrete. Computed tomography captures the microstructure of building materials and allows to study crack initiation and propagation. Manual segmentation of crack surfaces in large 3d images is not feasible. In this paper, automatic crack segmentation methods for 3d images are reviewed and compared. Classical image processing methods (edge detection filters, template matching, minimal path and region growing algorithms) and learning methods (convolutional neural networks, random forests) are considered and tested on semi-synthetic 3d images. Their performance strongly depends on parameter selection which should be adapted to the grayvalue distribution of the images and the geometric properties of the concrete. In general, the learning methods perform best, in particular for thin cracks and low grayvalue contrast.\\


\section{Introduction}
\label{sec:introduction}
As a building material, concrete needs to meet high standards regarding quality and sustainability. Therefore, investigating its performance under stress is an important field of research. Typically, the strength of concrete specimens is studied via stress tests that result in the formation of cracks. For instance, cracks in drill cores from vibrating beams without and with previous damage have been studied in \cite{weise15} while crack initiation and propagation in tensile tests have been analyzed in \cite{landis07,landis09}. 

The complex spatial morphology of cracks can be investigated in 3d using computed tomography (CT). However, segmenting crack surfaces in large CT images is challenging. Hence, the analysis is often limited to the visualization or analysis of virtual 2d sections \cite{paetsch19}. 

For crack detection and segmentation in 2d image data, a variety of methods are available: Grayvalue distribution based approaches \cite{rel:photo1, rel:photo2} analyze the histogram of the pixels' grayvalues, assuming that their distribution in the crack differs from that in the matrix. 
More advanced methods also incorporate mathematical morphology and shape analysis \cite{rel:morph2, rel:morph3, rel:morph4}. One example is template matching \cite{roseman}, where the similarity of a template with the local crack structure is evaluated. Hessian-based percolation \cite{yama} is a region-growing algorithm starting at preselected pixels and detecting further crack pixels based on a shape parameter.
Edge detection filters based on the local structure of an image can be used for crack segmentation, too. Such filters can be applied either on a single or on multiple scales \cite{rel:filters1, rel:filters2}. 
Finally, algorithms for computing minimal paths \cite{rel:minPaths1} can be applied based on the observation that cracks are connected structures that are darker than the matrix. Interpreting dark pixels as vertices of a network, they can be connected by minimal paths, e.g. via Dijkstra's algorithm \cite{rel:minPaths2}. 

Recently, deep and machine learning (ML) methods have become popular for solving image segmentation tasks. For example, supervised learning methods such as convolutional neural networks \cite{rel:ml1, segnet, Brox_2015} and random forests \cite{2drf2, 2drf} are able to segment 2d image data with high accuracy. ML is also used successfully to segment crack structures \cite{rel:ml2, bv-forum2018}. 

The survey \cite{rel:2d-review} provides a comprehensive overview 
of 2d crack segmentation methods.

In contrast to 2d, the examination of concrete specimen by means of 3d CT images is a rather novel approach and image acquisition is a challenging task. Hence, 3d image data of concrete featuring crack structures is still rather scarce. 

Adapting 2d methods to 3d images is difficult as there are fibrous (1d) and lamellar (2d) lower dimensional structures in 3d and discrete connectivity is far more complex in 3d than in 2d. Also, running times increase dramatically due to the large number of voxels such that algorithms for 3d images have to be much better optimized.

Nevertheless, filters, template matching, and region-growing approaches as well as minimal path algorithms have successfully been applied to 3d images of concrete \cite{paetsch11, paetsch12, rel:minPaths3}. 
Moreover, segmentation of thin structures in large 3d images is an important task in medical image processing, too. 
A typical task is the detection of tube-like structures such as blood vessels \cite{frangi}. With the sheet and Frangi filters, these methods are adapted to planar structures \cite{sheet_original, frangi_andre}, making them suitable for the segmentation of cracks.

Supervised learning methods can be extended to 3d image data as well. For instance, the 3d U-Net \cite{Brox_2016} is currently one of the most common convolutional neural networks for 3d image segmentation in the bio-medical field. In addition, random forests have been used for 3d segmentation \cite{ilastik}. 
For training, these methods require a set of images along with their segmentation. Usually, these are obtained by manual segmentation. 

However, the acquisition of training data for ML based segmentation of 
3d images is challenging: Classifying each voxel manually is highly subjective and, due to the amount of voxels, not feasible. 
A promising alternative is to generate synthetic or semi-synthetic images based on geometry models mimicking the structures of interest \cite{rel:chiara}. In this case, a ground truth is available which unambiguously distinguishes crack from non-crack voxels. 
Crack structures can be simulated based on physical models of concrete. To this end, the concrete's aggregates are often modelled as (Voronoi) tessellations and crack propagation is then simulated by finite element methods \cite{rel:sim1, rel:sim2}. 
Alternatively, cracks can be interpreted as stochastic structures, that is, as two-dimensional random processes. These can, for example, be simulated via two-dimensional fractional Brownian motions \cite{Addison_2000}. 

Here, we explore the possibilities and limitations of automated crack segmentation methods. A selection of methods including previously recommended 3d crack segmentation methods, successful 3d segmentation methods from other fields, and methods generalized from 2d are reviewed and compared. Both, classical image processing methods and ML approaches are considered.

Segmented real data that could be used for training the ML approaches is not available. Therefore, we propose a method for generating semi-synthetic 3d images. As the ground truth is known for these images, they can also be used to evaluate the crack detection methods:
We want to minimize the numbers of crack voxels that are missed and of falsely detected voxels in the background. For most methods, these are opposing objectives. Hence, they need to be discussed separately and under clearly defined constraints.

The aim of this paper is therefore threefold: First, we develop, adapt, and improve segmentation methods for 3d images of cracks in concrete. Second, we optimize their performance by choosing optimal parameters. 
Finally, we quantitatively compare the methods for a fixed set of semi-synthetic images containing cracks of varying shapes, sizes, and grayvalue distributions.

The paper is structured as follows: In Section \ref{ch:characteristics}, we summarize crack characteristics. Crack simulation based on the fractional Brownian motion is explained in Section \ref{ch:data-sim}. Section \ref{sectionMethods} summarizes the crack detection methods and describes the methods' parameters. In Section \ref{qualityMeasures} we introduce the performance measures. Parameter choice is discussed in Section \ref{sec:parameters}, and the performance of the optimally parametrized methods is compared in Section \ref{sec:comparison}. Finally, Section \ref{ch:conclusion} yields the conclusion and an outlook for future work.

\section{Crack and concrete characteristics}\label{ch:characteristics}

The heterogeneous structure of concrete and the complex morphology of cracks make the segmentation problem a challenging task. Concrete is a composite material usually consisting of aggregates, cement, air pores, and eventually reinforcement fibers. Cracks are thin, rather 2d substructures of the heterogeneous concrete matrix. The texture of cracks may vary depending on the composition of the concrete. Both, cracks with a smooth texture and rough cracks with spikes are possible. From the topological point of view, a crack is a connected object behaving locally like a 2d surface. When propagating, crack branches may emerge and cracks may intersect. 

The X-ray absorption of air is lower than that of the concrete  components. Thus, cracks appear darker than the surrounding background in the reconstructed CT images. As pores also contain air, their grayvalues in the CT images are similar to those of the cracks. This makes it impossible to distinguish pores and cracks solely based on the grayvalue distribution. 

\section{Generating semi-synthetic 3d images}
\label{ch:data-sim}

For simulating virtual CT images of concrete with cracks, we introduce a data simulation procedure which yields pairs of binary ground truth and semi-synthetic grayscale images. 

\begin{figure}[htbp]
	\centering
	\includegraphics[width=0.32\textwidth]{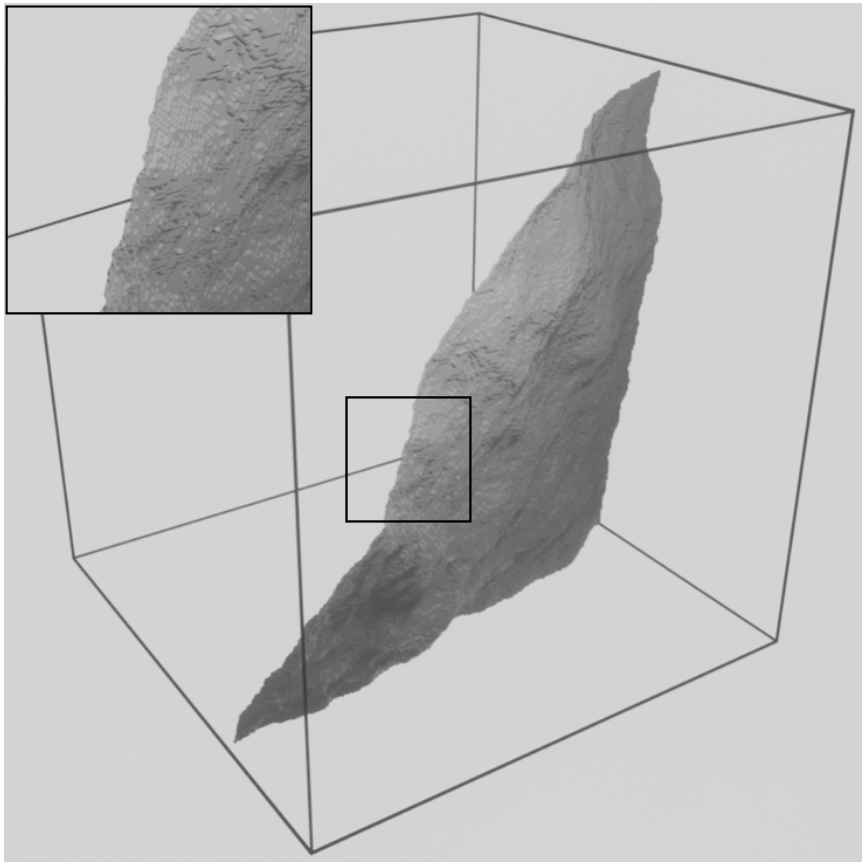}
	\includegraphics[width=0.32\textwidth]{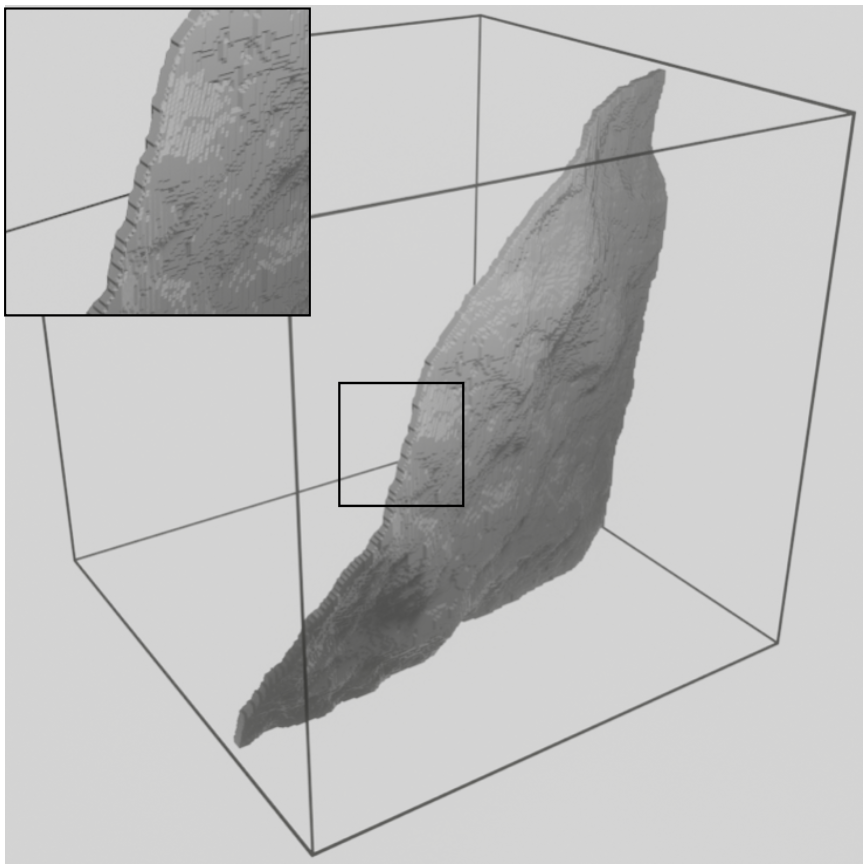}
	\includegraphics[width=0.32\textwidth]{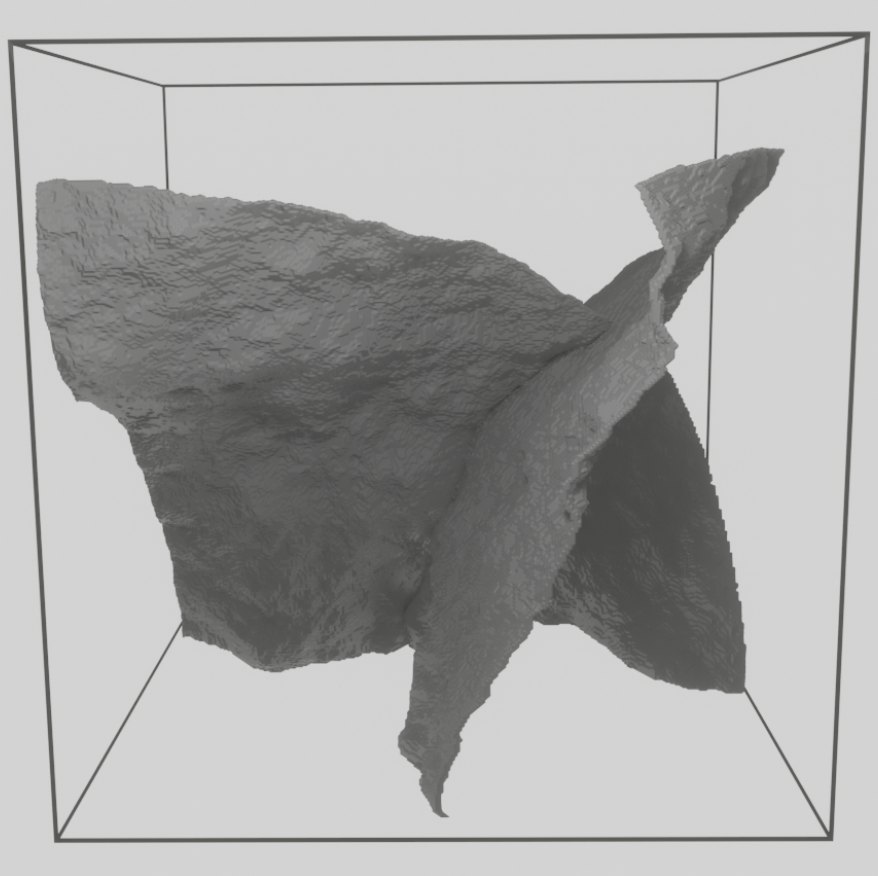}
	\caption{3d renderings of simulated crack surfaces for $H=0.99$ in an image of size $256^3$. Left: crack width 1, middle: crack width 5, right: two cracks of width three in orthogonal planes.}
	\label{fig:gt-w1-w3-w5}
\end{figure}

For the ground truth image of the crack, a fractional Brownian surface (fBS) with Hurst index $H \in [0,1]$ is simulated \cite{Addison_2000}. The larger the Hurst index, the smoother the resulting surface. The fBS is simulated using the MATLAB function \cite{fbm}. The output of the simulation is a realization $z$ of a random field on a two-dimensional array of size $2^{n} \times 2^{n}$, $n \in \mathbb{N}$. For discretization, the values $z(p,q)$ are scaled and rounded to integers in
$[-2^{n}/2 + 1, 2^{n}/2 - 1]$. In a 3d image of size $2^{n} \times 2^{n} \times 2^{n}$, voxels with coordinates $(p,q, z(p,q))$ are set to 1 (crack) while all others are set to 0 (background). The crack width can be varied by applying a dilation to the discretized fBS which represents a crack of width 1 voxel. Here, we dilate to widths of 3 and 5 voxels. Additionally, several simulated cracks can be combined in one image, see Figure~\ref{fig:gt-w1-w3-w5} for examples.  

\begin{figure}[h]
	\centering
	\includegraphics[width = \textwidth]{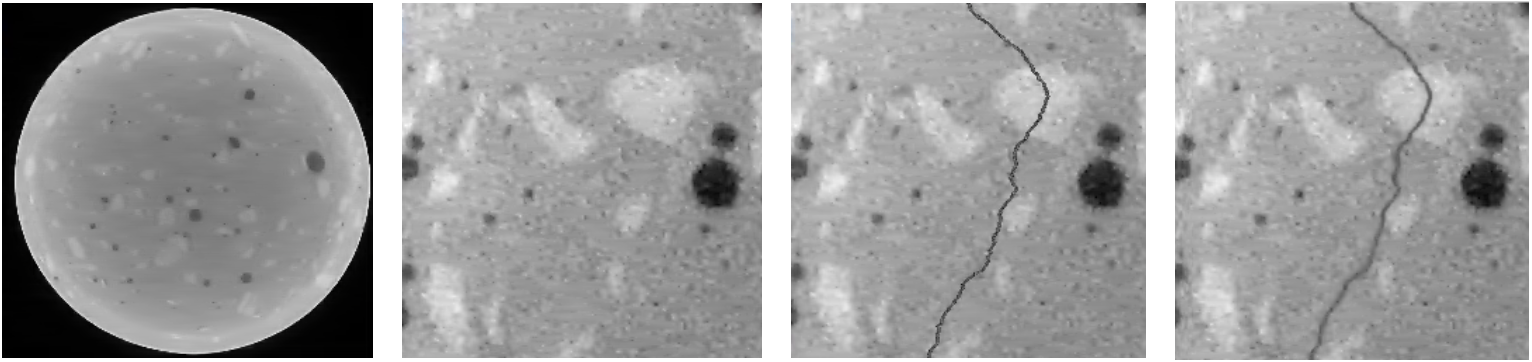}
	\caption{Generation of a semi-synthetic image. From left to right: 2d slice of a CT image of concrete, 2d slice of cropped image patch of size $256^3$, 2d slice of semi-synthetic image created by adding crack and background image, 2d slice of final semi-synthetic image (linear Gaussian filter applied to the dilated crack region).}
	\label{fig:gen-semi-synthetic}
\end{figure}

The grayscale images are then generated using 3d images of real concrete. From the available CT data, image patches of the same size as the ground truth images are extracted and multiplied with the inverse ground truth. In this way, the grayvalue in crack voxels is set to $0$ while the background remains unchanged. Next, grayvalues for the crack voxels are simulated by assuming that they are i.i.d. normally distributed. The mean value and standard deviation are estimated from the air pores in the CT images. 
Finally, we apply a linear Gaussian filter to the dilated crack region. This smoothes the transition between crack boundary and background and thus renders the image more realistic. The generation process is illustrated in Figure \ref{fig:gen-semi-synthetic}.

We create a data set of 60 semi-synthetic crack images of size $256^3$ voxels. It consists of 20 images per crack width 1, 3, and 5. Each of the three groups consists of eight images with one crack, six images with two cracks in parallel planes, and six images with two cracks in orthogonal planes. 

The background patches are extracted from four 3d CT images of concrete. The concrete samples represent the same concrete type consisting of aggregates, cement matrix, and air pores. The size of air pores and aggregates varies in each concrete specimen. Since the 3d CT images are sufficiently large, we can use different background patches for each image.

Out of the 60 images, nine (three per group) are used for training the learning methods. One image of each group is used as validation to tune the parameters of the proposed methods. Except for the training images, all images are used for evaluation of the methods.

\section{Methods for crack segmentation}
\label{sectionMethods}

In this section, we summarize the crack segmentation methods to be compared and evaluated. Throughout the section, $I:\mathbb{R}^3 \rightarrow \mathbb{R}$ denotes a 3d image.

\subsection{Hessian matrix}\label{sec:hessian}
Computing the Hessian matrix of an image is a common technique to detect edges and to describe the local structure voxelwise.

The second derivative of $I$ is given as

$$ \frac{\partial^2}{\partial x_i \partial x_j} I(p,\sigma) = \sigma I(p)*\frac{\partial^2}{\partial x_i \partial x_j}G(p,\sigma),$$
where $G:\mathbb{R}^3\times \mathbb{R}^+\rightarrow\mathbb{R}$ is the Gaussian kernel
with standard deviation $\sigma$.
Then, for each point $p\in\mathbb{R}^3$, the $3\times 3$ Hessian matrix is given as 
\begin{equation*}\label{hess::def}
H (p,\sigma) = \left(h_{i,j}\right)^3_{i,j=1}, \text{\hspace{0.7cm}} h_{i,j} = \frac{\partial^2}{\partial x_i \partial x_j} I(p,\sigma).
\end{equation*}

Here, $\sigma\in\mathbb{R}_{\geq 0.5}$ can be interpreted as a scaling parameter.

\begin{figure}[H]
	\centering
	\includegraphics[width = 0.23\textwidth]{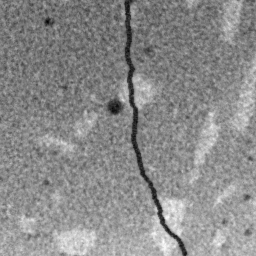}\
	\includegraphics[width = 0.23\textwidth]{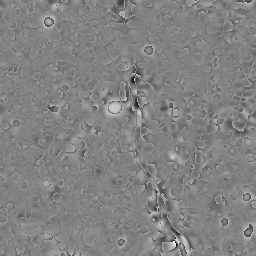}\
	\includegraphics[width = 0.23\textwidth]{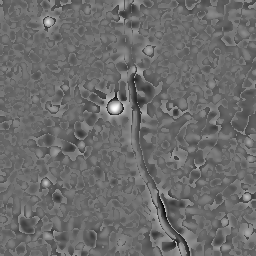}\
	\includegraphics[width = 0.23\textwidth]{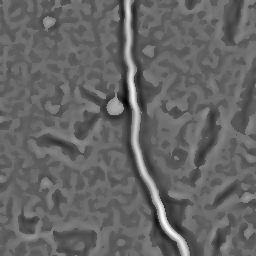}
	\caption{2d slices ($z=1$) of $256^3$ voxel images. From left to right: semi-synthetic input image and the eigenvalues $\lambda_1, \lambda_2, \lambda_3$ of the Hessian matrix of this input image.}
	\label{lambdas}
\end{figure}

Now let $|\lambda_1(p)|\leq|\lambda_2(p)|\leq|\lambda_3(p)|$ be the eigenvalues of $H(p,\sigma)$ at point $p$. As flat structures (cracks) have lower grayvalues than the image background, $H(p,\sigma)$ has small values in directions tangential to the cracks and high values in the orthogonal direction. That is, we expect that
$ 	\lambda_3(p)\gg\lambda_1(p) \simeq 0, \lambda_3(p)\gg\lambda_2(p) \simeq 0,
$ for crack voxels, see Figure~\ref{lambdas}.

\subsection{Classical image processing methods}

\subsubsection*{Sheet filter (SF)}
The sheet filter \cite{sheet_original} detects flat structures in an image based on a suitable composition of the Hessian eigenvalues. For $\delta\in\mathbb{R}_{> 0}$ and $\rho\in (0,1]$, let

$$g(\lambda_s,\lambda_t)=\left\{
\begin{array}{ll}
(1+\frac{\lambda_s(p)}{|\lambda_t(p)|})^{\delta} ,& \,  \, \lambda_s(p) \leq 0 \, , \, |\lambda_t(p)|\geq |\lambda_s(p)| \\

(1-\rho\frac{\lambda_s(p)}{|\lambda_t(p)|})^{\delta} ,& \,  \, \lambda_s(p) > 0 \, , \, \lambda_s(p)\leq |\lambda_t(p)|/\rho \\
0 ,& \,  \, \textrm{else.}
\end{array}
\right.$$

The parameter $\rho$ can be used to weight positive and negative values of $\lambda_1(p)$ and $\lambda_2(p)$ differently.
The sheet filter for voxel $p$ is defined as
$$S(p) = \left\{
\begin{array}{ll}
\lambda_3(p)\cdot g(\lambda_1(p),\lambda_3(p))\cdot  g(\lambda_2(p),\lambda_3(p)) ,& \, \lambda_3(p) > 0\\
0 ,& \, \textrm{else.} \\
\end{array}
\right.$$

The higher $S(p)$, the more likely the voxel $p$ belongs to a crack structure: The factor $\lambda_3(p)$ takes into account the assumption $\lambda_3(p)\gg 0$. Function $g$ measures the discrepancies between $\lambda_3(p)$ and $\lambda_1(p)$, $\lambda_2(p)$. The value of $g$ is close to 0 for small discrepancies and close to 1 for large ones. Thus, the former are penalized in the computation of $S(p)$.

To obtain a binary image, we apply a threshold $t_1\in\mathbb{R}$ on the resulting image $S$ and consider $p$ a crack voxel if $S(p)\geq t_1$.

\subsubsection*{Frangi filter (FF)}
The Frangi filter was originally developed to detect tube-like structures in medical images \cite{frangi} and can be adjusted to also detect plate-like structures \cite{frangi_andre}. Similar to the sheet filter, it is based on a comparison of the eigenvalues of the Hessian matrix. 
Let
\[Q_A\coloneqq Q_A(p,\sigma)
=\frac{|\lambda_2(p)|}{|\lambda_3(p)|},
\qquad
Q_B\coloneqq Q_B(p,\sigma)
=\frac{|\lambda_1(p)|}{\sqrt{|\lambda_2(p)||\lambda_3(p)|}}.
\]
For flat structures, both $Q_A$ and $Q_B$ should be close to 0. 
However, interpreting $Q_A(p,\sigma)$ and $Q_B(p,\sigma)$ is only meaningful if the voxel $p$ contains significant structure information. This is measured by the Frobenius norm of the Hessian,
$$R\coloneqq R(p,\sigma)=\sqrt{\sum_{i=1}^3 \lambda_i(p)^2}.$$

Appropriately putting together the terms for $Q_A,Q_B$, and $R$, we define 
$$E(p)=\left\{
\begin{array}{ll}
\exp(-Q^2_A/\alpha)\exp(-Q^2_B/\beta)(1-\exp(-R^2/\eta)),&\lambda_3(p) > 0 \, , \, \lambda_2(p)\neq 0 \\

\exp(-Q^2_A/\alpha)(1-\exp(-R^2/\eta)),&\lambda_3(p) > 0 \, , \, \lambda_2(p)=0 \\
0,&\textrm{else}
\end{array}
\right.$$

for each voxel $p\in I$, where $\alpha,\beta>0$ are scaling parameters and $\eta\coloneqq \eta(p,\sigma) = \max_{p} R(p,\sigma).$ We write $E(p)=E(p,\sigma)$ to emphasize the dependence on the scale parameter $\sigma$. Let $0. 5\le \sigma_{min} \le \sigma_{max}\in\mathbb{R}$. We define the Frangi filter as
$$F(p)=\max_{\sigma_{min}\leq \sigma\leq \sigma_{max}} E(p,\sigma), \quad p \in I. $$ We observe that $F(p)\in[0,1)$. The higher $F(p)$, the more likely the voxel $p$ belongs to a crack structure.
Again, we apply a threshold $t_2\in[0,1)$ on the resulting image $F$ and consider a point $p$ a crack voxel if $F(p)\geq t_2$.

\subsubsection*{Template matching (TM)}
The basic idea of template matching is to investigate how well a given pattern $T$ fits the image content in the neighbourhood of voxel $p = \left(r,s,l\right)$. It was adapted to crack detection using a plate-like template in \cite{paetsch11}. 

\begin{figure}[H]
\centering
\includegraphics[width = 0.44\textwidth]{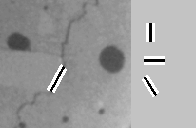}
\includegraphics[width = 0.4518\textwidth]{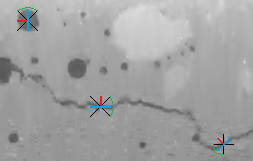}
\caption{Left image: template matching: best fitting template and selected template rotations. Note that the template is inverted matching the originally dark crack here to keep consistency of all concrete figures. Right image: illustration of adaptive morphology for plane filtering: normal (red), initial plane (blue), search cone (black and green).}
\label{fig:template-matching}
\end{figure}

The template matching is performed on the inverted image such that crack voxels have higher grayvalues than the background. Hence, the base template $T$ is chosen as an image of size $\left(2N+1\right)\times \left(2N+1\right)\times(b+c+b)$ consisting of two background layers of thickness $b$ with grayvalue 0 enclosing a central crack layer of thickness $c$ with grayvalue 1. 
An example of a base template $T$ which models cracks of width $1$ ($b=c=1$) is given as
\begin{equation*}
T\left(:,i,:\right) = 
\begin{bmatrix}
0 & 0 & ... & 0 & 0\\
1 & 1 & ... & 1 & 1\\
0 & 0 & ... & 0 & 0\\
\end{bmatrix}, \quad i = 1,\ldots,\left(2N+1\right).
\end{equation*}
We consider rotated copies $T_{\theta}$ of the template with normals pointing to a prescribed set of directions $\theta$ regularly distributed on the unit sphere. For each voxel $p$ we determine the rotation resulting in the best fit of the template to the local grayvalue structure, see Figure \ref{fig:template-matching}. 
The goodness of fit is measured by the cross-correlation coefficient
\begin{equation*}
C_{\theta}(p)= C_{\theta}\left(r,s,l\right) = \frac{1}{N} \sum_{\left(i,j,k\right) \in T_{\theta}}{\frac{\left(I\left(r+i,s+j,l+k\right) - \overline{I} \right)\left(T_{\theta}\left(i,j,k\right) - \overline{T} \right)}{\sigma_I \sigma_T}},
\end{equation*}
where $\overline{I}$ and $\overline{T}$ are the means of grayvalues of $I$ in the voxels covered by $T_{\theta}$ and $T$, respectively. $\sigma_I$ and $\sigma_T$ are the corresponding standard deviations of grayvalues.

The template size parameters $N$, $b$, and $c$ should be chosen such that the template fits the size and thickness of the crack. For the orientation $\theta$ we use  
a regular sampling of points on the unit sphere, following \cite{AltendorfThesis}.
It requires the choice of a discretization parameter $n$. For given $n$, the sample consists of roughly $n (n/4+1)$ points.
A finer discretization of the sphere allows for a closer fit of the template to the cracks but also results in longer computation. 

Finally, a threshold $t_4 \in [0,1]$ is chosen. Voxels $p$ where $ \max_{\theta} C_{\theta}(p) \ge t_4$ are labelled as crack voxels.  

\subsubsection*{Adaptive plane morphology (AM)}
Adaptive plane morphology is based on filtering using plate-like structuring elements. Here, we use the median filter but other filters can also be applied. To avoid searching the whole orientation space as done in template matching, this approach uses local information to estimate the locally dominating orientation \cite{morpho-hess}. 
The structuring element is rotated to a prescribed set of directions in the cone around this orientation searching for the best filter response \cite{barisin21}. This results in significant reduction of computation time. The basic idea is illustrated in Figure~\ref{fig:template-matching}. 

The parameters of the method include the half-length $N$ of the plate structuring element of size $\left(2N+1\right)\times\left(2N+1\right)\times1$,
the discretization parameter $n$ of the unit sphere as in template matching,
and the half opening angle $\delta_{max}$ of the search cone. 
The initial local orientation estimate in a voxel $p$ is supposed to be the local normal direction of the crack plane. It is estimated by the eigenvector belonging to the largest eigenvalue of the Hessian matrix $H(p, \sigma)$. 

Adaptive filtering as described above smoothes the image and emphasizes planar structures. In a second step, dark image regions should be classified into cracks (2d structures) and pores (truly 3d objects). For this purpose, we apply a median filter on the line in the normal direction of the best fitting plate following the idea from \cite{Sazak2019}. A large difference of the two filter responses then indicates crack voxels. 
The threshold value $t_5$ is computed as 
$
t_5 = \mu_{IM} + k \sigma_{IM},
$
where $\mu_{IM}$ and $\sigma_{IM}$ are the mean value and the standard deviation of the difference of the filter responses.
The parameter $k$ balances between coverage of cracks and misclassification of noise as cracks.

\subsubsection*{Minimal paths (MP)}
This algorithm generalizes the 2d method described in \cite{avila_2014}. Details can be found in \cite{rel:minPaths3}. 
The procedure is based on modeling the image as a 3d vertex-weighted, directed graph. The voxels are the nodes weighted by their grayvalues. The set of directed edges is defined by a voxel neighborhood associated with a direction. Each neighborhood consists of nine discrete directions in the 26-neighborhood voxel grid. For instance, the direction \textit{up} connects each voxel with its nine neighbors in the plane above it.
In total, 18 neighborhoods centered in the six coordinate directions (with positive and negative sign) and the twelve plane diagonal directions are considered.  

\begin{figure}[h]
\centering
\includegraphics[width = 0.8\textwidth]{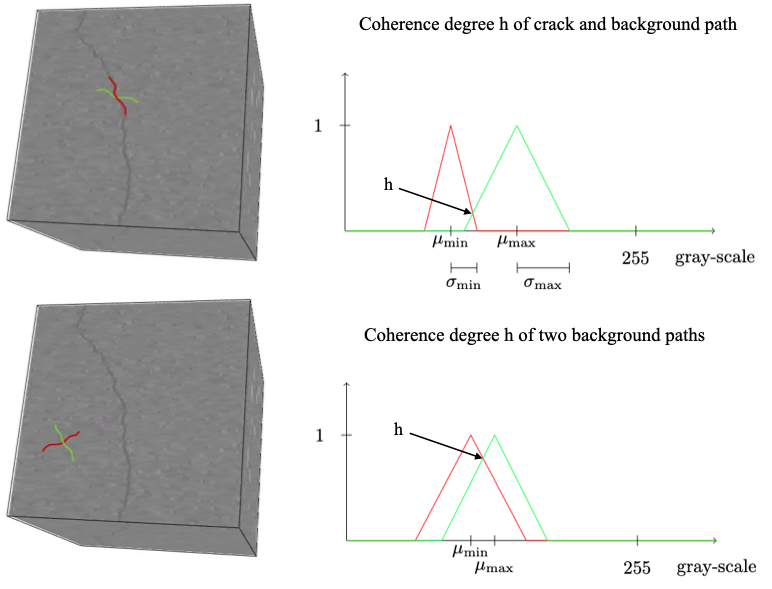}
\caption{Voxelwise classification by measure of coherence.}
\label{fig:coherence-measure}
\end{figure}

This results in 18 different 3d graphs for one image. In each graph and for each vertex $p$, minimal paths are computed by a simple percolation algorithm. Starting in $p$, the neighbor with the minimal weight is successively added to the path until a predefined path length $\ell$ is reached. A \textit{local minimal path} is then obtained by merging the two paths of opposite directions. 
This procedure results in nine local minimal paths passing a voxel $p$. The decision whether $p$ belongs to a crack is taken by contrasting the grayvalue distributions in the two paths with minimal and maximal mean grayvalue. To do so, the coherence measure $h$ introduced in \cite{avila_2014} is applied. The idea is illustrated in Figure \ref{fig:coherence-measure}. The measure $h$ has values between zero and one, where a value close to zero indicates that the voxel belongs to a crack. 
Hence, applying a threshold $t_3 \in [0,1]$ yields a segmented crack image. 

\subsubsection*{Hessian-based percolation (HP)}
A region growing algorithm for crack detection in 2d images of concrete is described in \cite{yama}. Starting from each pixel of a 2d image $I$, a connected set of pixels $P$ is computed iteratively based on a dynamic threshold $t$. A starting pixel $p$ is labelled as crack pixel if the shape of $P$ matches a certain criterion. More precisely, the algorithm consists of the following steps:
\begin{enumerate}
\item Initialize $P=\{p\}$ and $t=I(p)+\varepsilon$ for some $\varepsilon\in\mathbb{R}$.
\item Add each neighbor $q$ of $P$ with $I(q)\leq t$ to $P$.
\item Update $t$ by setting $t=\max\left(\max_{q\in P} I(q),t\right)+\varepsilon.$
\item Repeat steps 2 and 3 until $P$ comes in contact with the boundary of a window of size $(2W+1)\times (2W+1)$ pixels with center $p$.
\item Pixel $p$ belongs to the crack if 
$F_{2d}\coloneqq\frac{4\text{area}(P)}{\pi (\text{diam}(P))^2}\leq f$ for given $f\in[0,1]$
\end{enumerate}
The parameter $\varepsilon$ can be chosen freely and is used to speed up the algorithm. The fraction $F_{2d}\in[0,1]$ can be interpreted as the circularity of $P$. $F_{2d}$ being close to 0 indicates an elongated shape of $P$ such that the starting pixel is likely to be a crack pixel.

Straightforward adaption of the algorithm to 3d images results in an enormous runtime. Therefore, the following modifications are suggested in \cite{paetsch11}:  

\begin{enumerate}
\item 
The percolation process is started only from a preselected set $H$ of crack candidate voxels instead of all voxels. This preselection can be obtained, for example, by applying the sheet filter or the Frangi filter.
\item Instead of the circularity $F_{2d}$, compute 
$F_{3d}\coloneqq|P\cap H|/|P|\in[0,1]$ where $|\cdot|$ denotes the number of voxels in the respective set.
\item If $F_{3d}\geq f$, \textit{every} voxel in $P$ is considered a crack voxel.
\item Some voxels may be labelled as crack voxels multiple times.  Hence, we introduce a parameter $\tau$ below which rarely detected voxels are discarded.
\end{enumerate}

\begin{figure}[H]
\centering
\includegraphics[width = 0.3\textwidth]{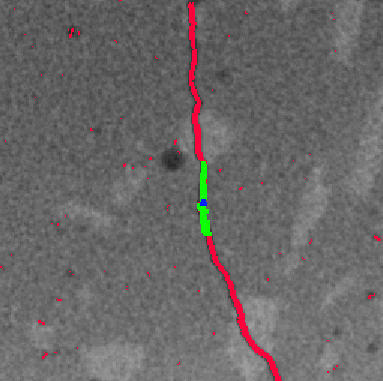}
\includegraphics[width = 0.3\textwidth]{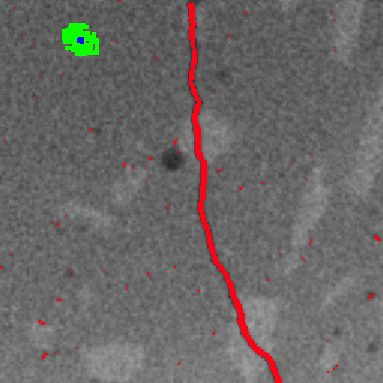}
\caption{Slice $z=100$ of a semi-synthetic image of size $256^3$. Red: set $H$ of voxels preselected by the Frangi filter, blue: starting voxel, green: percolated set $P$. Left: set $P$ is labelled as crack as $F_{3d}$ is high, right: $F_{3d}$ is not labelled as crack as $F_{3d}$ is comparatively low.}
\label{algperc}
\end{figure}

The algorithm is visualized in Figure \ref{algperc}. Note that a good preselection of voxels is necessary for the algorithm to perform well. 

\subsection{Supervised learning methods}
\subsubsection*{Random forests (RF)}
Tree-based methods can be used to segment 2d and 3d image data. In particular, random forests have been proven to be an appropriate tool to segment 2d images of concrete, see \cite{2drf} and \cite{2drf2}. A generalization to 3d images is straightforward.

A random forest is a collection of decision trees, each trained on several features. The features are the grayvalues of voxels of the original image and of several image transforms. Note that each tree is trained on a different training data set obtained by using the bootstrap procedure on the original training data: For every tree, the feature vectors are drawn randomly and with replacement until the size of the original training data set is reached.  

The output of a decision tree is a classification of the voxel as crack (1) or no crack (0). As the classifications of the single trees may differ, the final prediction is obtained via majority voting.

The main parameters of the method are the depth of the trees $d_{dt}$ and the maximum number of trees $n_{dt}$. Additionally, the parameters of the image transforms used for constructing the features have to be chosen.

\subsubsection*{Deep learning: 3d U-Net (NN)}
The 3d U-Net was originally introduced for biomedical image segmentation tasks \cite{Brox_2016}. It is a convolutional neural network with characteristic u-shape resulting from a composition into encoder and decoder. The encoder is responsible for capturing the image context. Then, the decoder expands the downsampled feature map back to the original image size. In order to prevent overfitting, we add a dropout layer with a dropout probability of $0.5$ at the bottleneck between decoder and encoder. The output of the network is an image of the same size as the input image containing values in $[0,1]$, the probabilities for each voxel of belonging to a crack. In order to obtain a binary image, a threshold $t_6$ in $[0,1]$ is applied.

\section{Quality measures}\label{qualityMeasures}

In the following, we use the notation \textit{tp} (true positive) for the number of correctly predicted crack voxels, \textit{tn} (true negative) for the number of correctly predicted background voxels, \textit{fp} (false positive) for the number of falsely predicted crack voxels, and \textit{fn} (false negative) for the number of falsely predicted background voxels.

Precision (P), recall (R) and F1-score (F1) are defined as 
$$P = tp/(tp+fp),\quad R = tp/(tp+fn),\quad F1 = 2PR/(P + R),$$
where the F1-score is the weighted average of precision and recall.

Precision provides information on how exact a positive result is, i.e. what proportion of voxels classified as positive are indeed positive. Recall measures the fraction of positive voxels that are classified correctly. In general, precision is a good measure when \textit{fp} should be penalized more while recall puts more weight on \textit{fn}. The F1-score is used when a balance between precision and recall is pursued. In particular, it is a suitable overall measure when dealing with class imbalance. Note that alternative measures, as for instance the accuracy, may be less meaningful due to the high percentage of background voxels.

For segmentation tasks, it is not unusual to introduce a certain voxel tolerance $tol$. 
That means: A true crack voxel is considered \textit{tp} in the output if its distance to the positives is less or equal \textit{tol}. Otherwise, it is considered \textit{fn}. 
A predicted crack voxel is only counted as \textit{fp} if its distance to the true crack voxels is larger than \textit{tol}.

\section{Choosing optimal parameters}
\label{sec:parameters}
Most of the methods introduced in Section~\ref{sectionMethods} use parameters that can be tuned to the data set to be examined. Due to the availability of a ground truth, we can optimize the parameters for our semi-synthetic images.
We tune the parameter values based on one fixed image per crack width ($1$, $3$, and $5$). The remaining images are then segmented using these tuned parameter values to assess the robustness of the methods w.r.t. parameter selection. 
For methods with three or more parameters, we choose a `grid-search' approach: Fixing all but one parameter, we compute the results for a range of values for the parameter that is not fixed. We choose the value that yields the best result and apply this strategy to the remaining parameters.

Parameter selection requires a trade-off between optimizing precision and optimizing recall. Optimizing recall penalizes false negatives and thus tends to give more false positives. This is advantageous if the crack does not have to be segmented exactly and a rough region in which the crack is located is required. On the contrary, optimizing precision
penalizes misclassification of concrete voxels as cracks (i.e. false positives), and as a consequence might miss some crack voxels. Hence, we decided to tune parameter values separately for precision and recall. Yet, when doing so, we have to choose suitable constraints to ensure that the results are still meaningful. 

For the deep and machine learning methods, we do not distinguish between the two objectives of optimizing precision and recall. The learning methods are able to unite both objectives due to their high performance.

In the following, we summarize our observations regarding the choice of adequate parameter values. Due to space restriction, we only report the results for crack width 3 voxels. 
The values for other crack widths can be found in Appendix \ref{app:Para}.

\subsection*{Hessian matrix}

The parameter $\sigma$ should be set to half of the width of the structure to be detected \cite{sheet_original, frangi}. The choice is confirmed by the grid-search for both the sheet filter and the Frangi filter. Slight deviations of up to +/-0.3 from this recommended value affect the results only marginally, larger deviations are not recommended. In adaptive plane morphology, $\sigma$ only plays a minor role as the search procedure compensates for rough scale selection.

\subsection*{Choosing thresholds}
Almost all methods use a threshold as a final step to transform their specific outputs to a binary image. 
In general, a smaller threshold gives better recall values while a larger threshold yields a better precision. Hence, a threshold that balances the two objectives should be chosen. 
The range of the methods' output voxel values may depend on properties of the particular image such as brightness or contrast. Hence, we do not expect our recommendations below to be generally applicable. However, computing a threshold on an image is comparably fast, such that this last step can easily be done adaptively.

\subsubsection*{Sheet filter (SF)}
\begin{figure}[h]
	\centering
	\includegraphics[width = 0.9\textwidth]{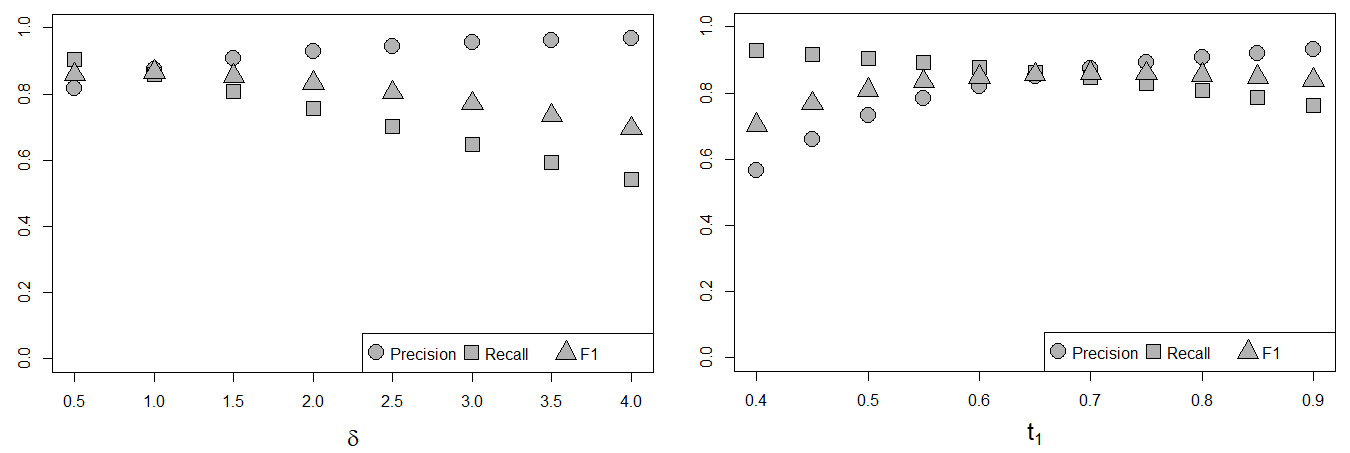}
	\caption{Sheet filter applied to the test image with crack width 3. Precision, recall, and F1-score for $\rho=1$, $\delta\in[0.5,4]$, $t_1=0.8$, $\sigma = 1.5$ (left) and for $\rho=1$, $\delta=1.5$, $t_1\in[0.4,0.9]$, $\sigma = 1.5$ (right).}
	\label{fig:sheet-parameters}
\end{figure}

Figure~\ref{fig:sheet-parameters} shows the influence of the parameters $\delta$ and $t_1$ on the results for crack width 3. Based on the plot, we choose $\delta =1.5$ to optimize precision and $\delta=1$ to optimize recall. The smaller the crack width, the larger one should choose $\delta$. The threshold $t_1$ can be chosen in the range $[0.6, 0.8]$.

The parameter $\rho$ only has a minor influence on the results. We set $\rho=1$ when optimizing precision for all crack widths. When optimizing recall, $\rho$ should be decreased for increasing crack widths. 
In total, we found the sheet filter to be comparatively robust with respect to the chosen parameter configuration.

\subsubsection*{Frangi filter (FF)}
\begin{figure}[h]
	\centering
	\includegraphics[width = 0.9\textwidth]{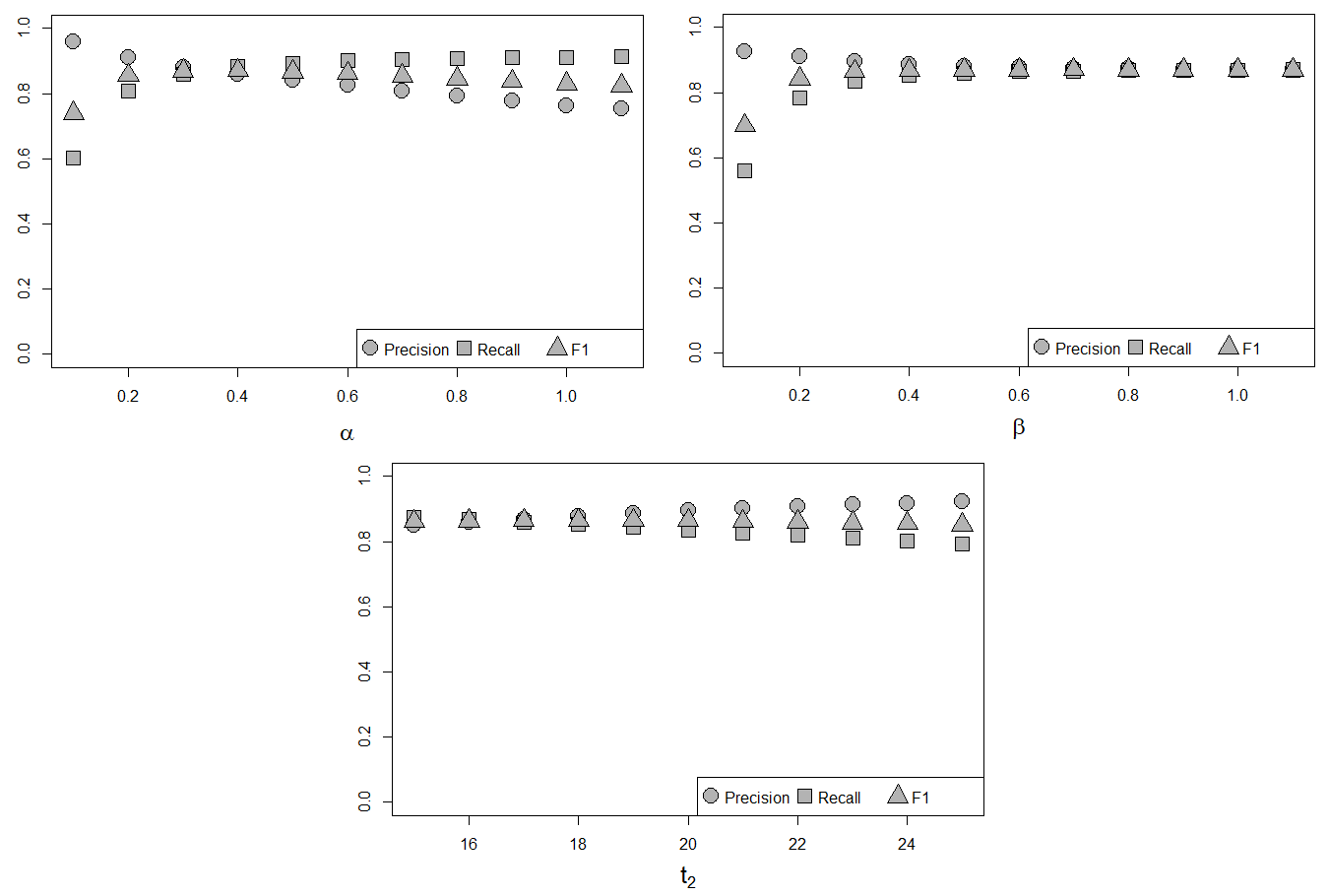}
	\caption{Precision, recall and F1-score for different parameter configurations of the Frangi filter and a crack of width $3$. Top left: $\alpha\in[0.1,1.1]$, $\beta=0.5$, $t_2=20$, top right: $\alpha=0.3$, $\beta\in[0.1,1.1]$, $t_2=20$, bottom: $\alpha=0.3$, $\beta=0.3$, $t_2\in[15,25]$.}
	\label{fig:frangi-parameters}
\end{figure}

As we only consider fixed crack widths, we set $\sigma_{\textrm{min}} = \sigma_{\textrm{max}} = \sigma$ and choose $\sigma$ as half the crack width.  
W.r.t. the choice of the remaining parameters, the filter shows a very robust behavior, see Figure \ref{fig:frangi-parameters}. $\alpha$ should be chosen slightly larger when optimizing recall compared to optimizing precision. For $\beta$ any choice larger 0.3 yields a stable result. For cracks of width 1, $\beta$ is increased slightly.
To improve the robustness of the Frangi filter w.r.t the choice of $t_2$, we normalize the filtered image to an 8-bit grayscale image. The threshold $t_2$ can then be chosen in the range $[18,28]$ for all crack widths.

\subsubsection*{Template matching (TM)}
\begin{figure}[h]
	\centering
	\includegraphics[width = 0.9\textwidth]{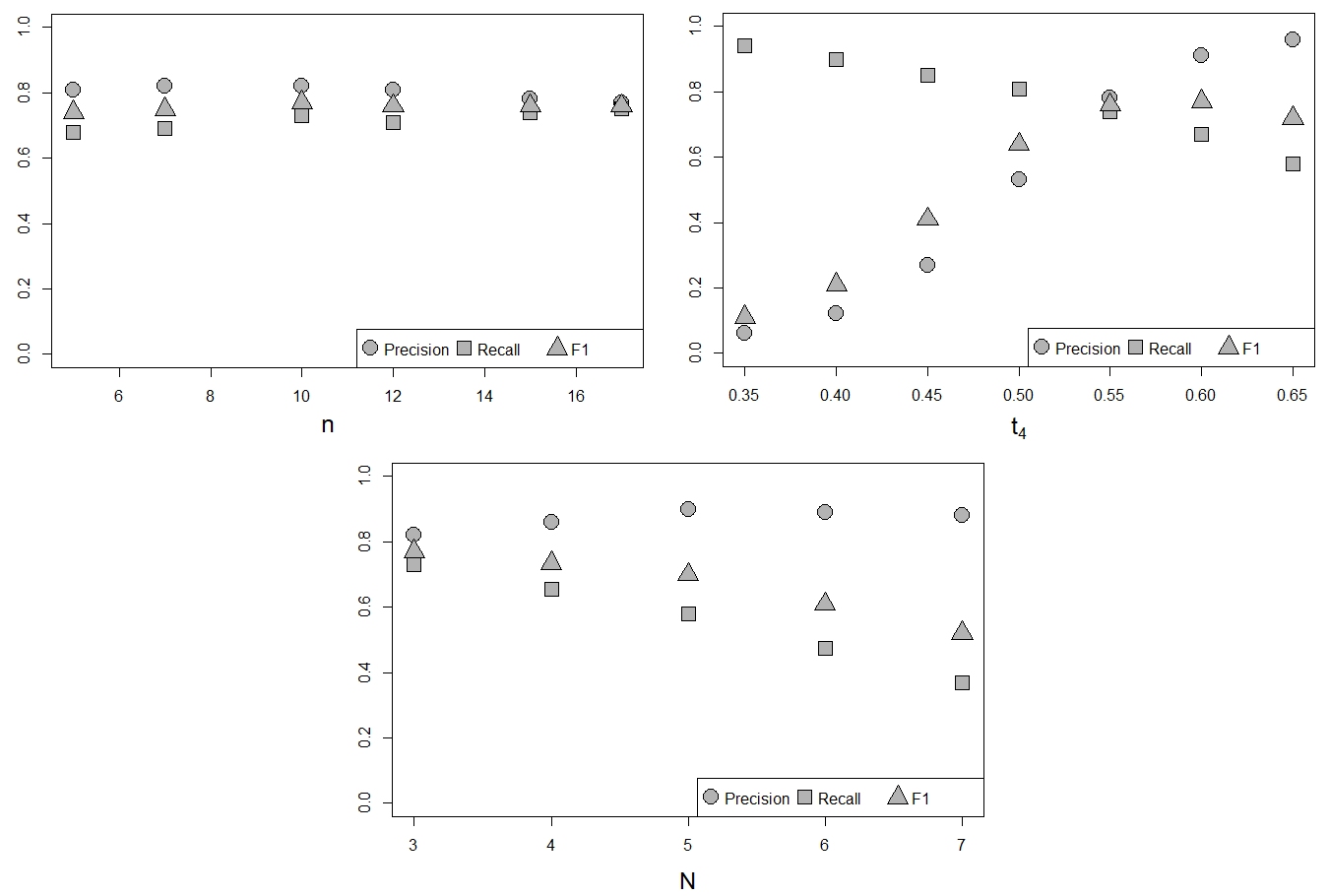}
	\caption{Precision, recall and F1-score for template matching on crack width $3$. Remaining parameters: $N=3, t_4 = 0.55, b=c=3$ (top left), $N=3, n = 15, b=c=3$ (top right), $n = 10, t_4 = 0.55, b = c = 3$ (bottom).}
	\label{fig:template-parameters}
\end{figure}

The width parameters $b$ and $c$ of the template are chosen for each crack width separately. For crack width $3$, we set $b=c=3$. The results of tuning the remaining parameters $n$, $N$, and $t_4$ for crack width $3$ are shown in Figure \ref{fig:template-parameters}. 

For the discretization parameter $n$, we expect that a fine discretization enables better fitting of the template to the crack structure. Nevertheless, results are quite robust to the choice of $n$. The plate length $N$ should not be chosen too large, as that decreases the quality of the fit due to cracks being curved structures. Finally, choosing $t_4$ around $0.6$ achieves optimal precision while $t_4$ around $0.4$ yields optimal recall. Here, similar parameters can be chosen for all three crack widths.

\subsubsection*{Adaptive plane morphology (AM)}
\begin{figure}[h]
	\centering
	\includegraphics[width = 0.9\textwidth]{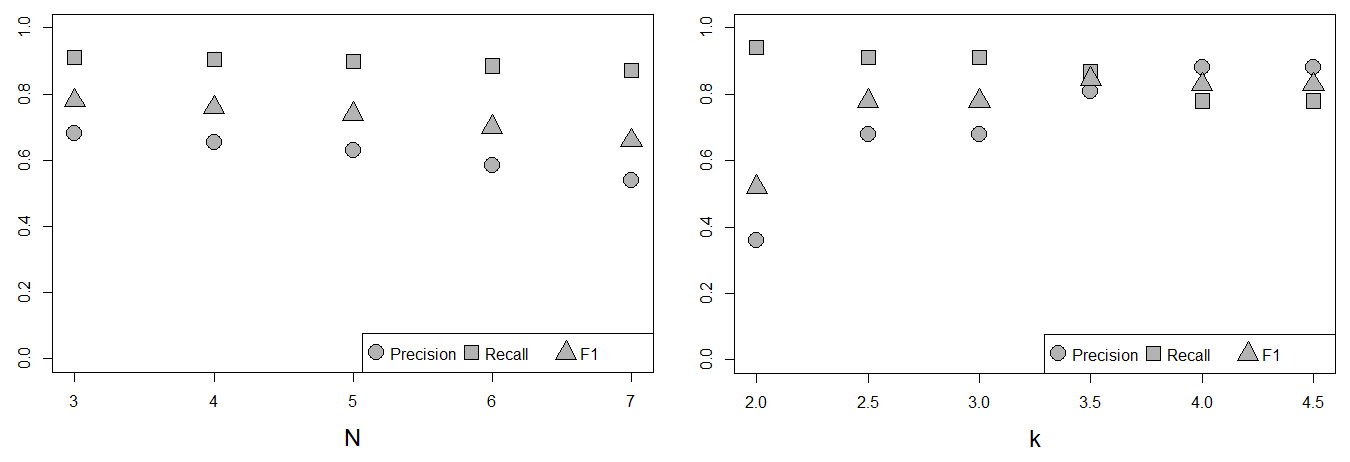}
	\caption{Precision, recall, and F1-score for adaptive morphology on crack width $3$. Remaining parameters: $k = 3, \delta_{max} = 0.5, n = 20$ (left), $N = 3, \delta_{max} = 0.5, n = 20$ (right).}
	\label{fig:adaptive-parameters}
\end{figure}

Some parameters of this method overlap with the template matching approach: The half-length of the plate $N$ and the discretization parameter $n$ in this method have a similar role as above. Also, here we find that the plate should ideally be around $7$ pixels long ($N=3$), see Figure \ref{fig:adaptive-parameters}.
The cone width parameter $\delta_{max}$ is fixed to $0.5$ which yields enough variability in the searching procedure but avoids checking too many orientations.

Since our output values do not scale in the interval $\left[0,1 \right]$, we follow a different type of thresholding than in template matching. Here, the parameter $k$ controls the number of standard deviations $\sigma_{IM}$ 
that will be taken into account. This thresholding approach is motivated by the well-known `three-sigma' rule for the Gaussian distribution.
We have found that a choice of $k$ between $4$ and $4.5$ results in an optimal precision. An optimal recall is achieved for $k$ between $2$ and $2.5$. Parameter recommendations are valid for all three crack widths.  
As outlined above, the choice of the parameter $\sigma$ of the Hessian matrix only plays a minor role. We chose $\sigma =1$ for crack width 3 and 5 and $\sigma =0$ for crack width 1.

\subsubsection*{Minimal paths (MP)}
\begin{figure}[h]
	\centering
	\includegraphics[width = 0.48\textwidth]{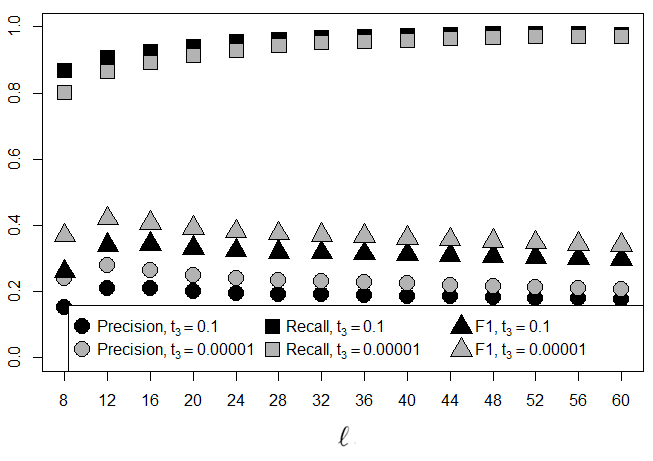}
	\caption{Precision, recall, and F1-score for different parameter values of the minimal paths algorithm applied to crack width 3.}
	\label{fig:minpath-parameters}
\end{figure}

The proposed algorithm has two parameters, the path length $\ell$ and a threshold $t_3$. As we have only two parameters, we deviate from our initial strategy and vary both parameters at the same time.
The results for crack width $3$ and the maximal and minimal value of $t_3$ are shown in 
Figure~\ref{fig:minpath-parameters}.

In contrast to the other methods, the recall scores are always significantly better than the precision scores. 

For large $\ell$, all three scores become practically constant. Rather low values of $\ell$ yield a better precision. A good recall is obtained with larger values of $\ell$. 

The behaviour for crack width $1$ and $5$ is very similar. Thus, we choose $\ell = 12$ to optimize precision and $\ell= 48$ to optimize recall for all three crack widths. As small threshold values yield a better precision, we fix $t_3=10^{-4}$ in all cases. For recall and for a large $\ell$, the threshold values have almost no influence (in the considered range).  

\subsubsection*{Hessian-based percolation (HP)}

\begin{figure}[h]
	\centering
	\includegraphics[width = 0.9\textwidth]{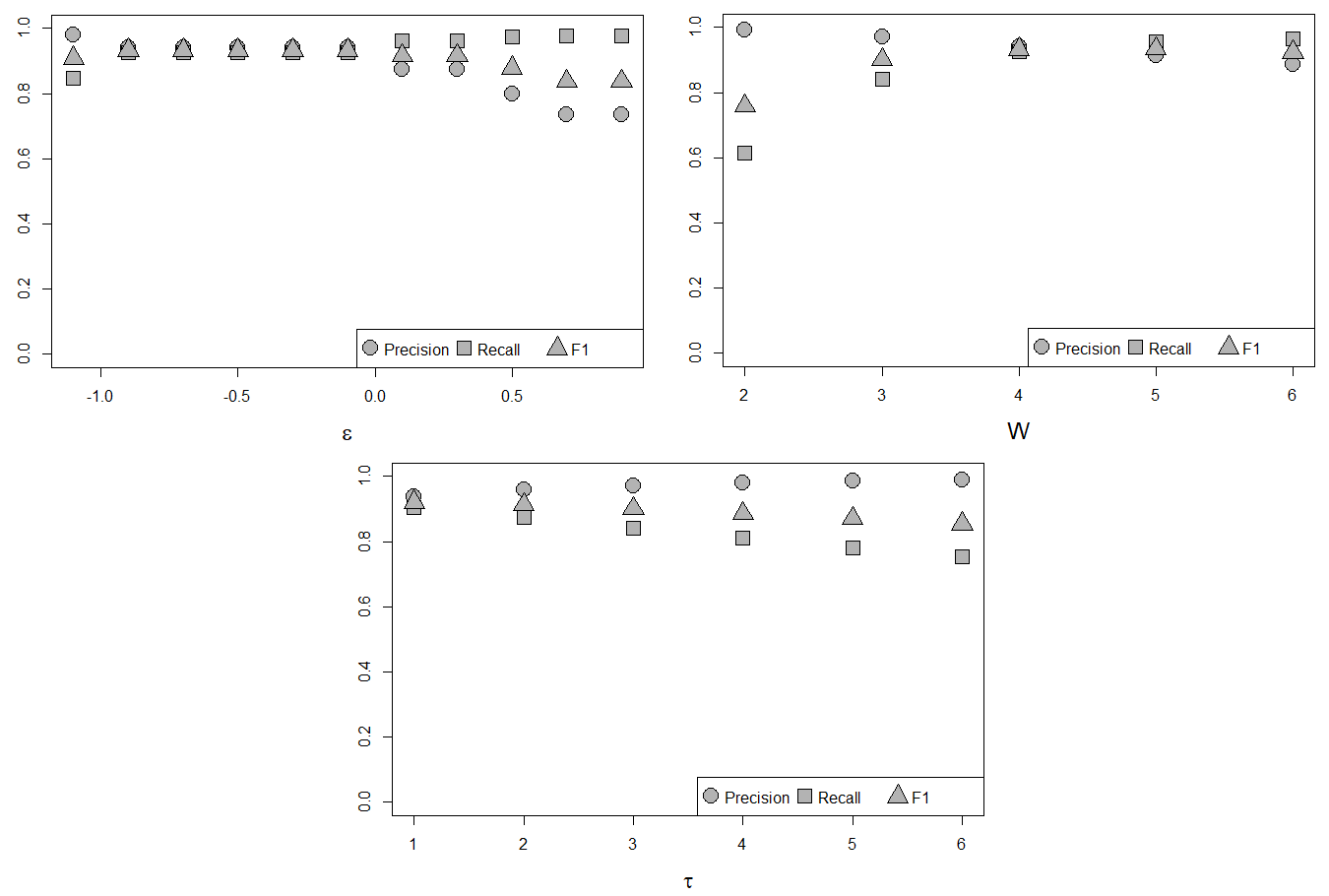}
	\caption{Precision, recall, and F1-score for different parameter configurations of the percolation algorithm for crack width $3$. Top left: $\varepsilon\in[-1.1,0.9]$, $W=4$, $f=0.6$, $\tau=3$, top right: $\varepsilon=-0.5$, $W\in[2,6]$, $f=0.6$, $\tau=3$, bottom: $\varepsilon=-0.5$, $W=4$, $f=0.6$, $\tau\in[1,6]$.}
	\label{fig:perc-parameters}
\end{figure}

Comparing Figures~\ref{fig:sheet-parameters} and \ref{fig:frangi-parameters}, we see that the Frangi filter yields higher recall values than the sheet filter. The percolation algorithm is able to remove $fp$ rather accurately. Therefore, we choose the Frangi filter to compute the preselected set of voxels. Its parameters are chosen as discussed above. 
The influence of the parameters $\varepsilon$, $W$, and $\tau$ for the input image with crack width 3 is shown in Figure~\ref{fig:perc-parameters}.

A choice of $\varepsilon\in[-0.9,-0.1]$ yields almost constant and reasonable results. Lower values increase precision and higher values increase recall.
Furthermore, we choose $W\in\{3,4,5\}$. For larger $W$ we obtain a higher recall and a lower precision. The parameter $f$ can be chosen as $f\in[0.60,0.87]$. Lower values yield rather constant results. Higher values can be used to increase precision slightly with a comparatively large decrease of recall. The parameter $\tau$ can be interpreted as a threshold. We found that increasing $\tau$ leads to an increase of precision and a decrease of recall. In practice, a reasonable compromise is achieved for $\tau\in[1,5]$. The parameter choices discussed apply for all crack widths tested. 

\subsection*{Supervised learning methods}
Besides the model specific (hyper-)parameters, the supervised learning methods require a training step. The choice of training data has a major influence on the result. 
We train on semi-synthetic training data generated as described in Section~\ref{ch:data-sim}. We use the same fixed training data for the random forest and the U-Net. We train separate models for the three crack widths ($1$, $3$, and $5$). In all cases, the training set consists of three images of size $256^3$ (an image with one crack, an image with two parallel cracks, and an image with two orthogonal cracks). 

\subsubsection*{Random forest (RF)}
The quality of the results of a random forest segmentation depends on its training data, the parameter choice, and the selection of image features. 
The features resulting from image transforms given in Table \ref{rf:features} have proven to train well-performing random forest classifiers for 2d and 3d image segmentation in different contexts \cite{ilastik}. Adjusting the parameters, we were able to obtain adequate results for the problem of crack detection as well. The definitions of the image transforms are given in Appendix \ref{app:RF}.

When tuning $n_{dt}$ and $d_{dt}$, we face a trade-off between goodness and run-time. Choosing $n_{dt}=100$ and $d_{dt}=50$ yields reasonable results without an exceptional computational effort. Increasing those values only results in a marginal increase of the goodness. Due to the large amount of features, decreasing $d_{dt}$ and $n_{dt}$ decreases the robustness of the random forest. 

\begin{table}[h]
	\centering
	\begin{tabular}{|l|l|l|}
		\hline
		Image transform & \multicolumn{2}{c|}{Parameters} \\ \hline
		Gaussian & $\sigma$ & 0.5, 0.75, 1.0, 1.5, 2.5, 3.5, 5.0\\ \hline 
		Laplacian of Gaussian &  $\sigma$ & 0.5, 1.0, 1.5, 2.5, 3.5, 5.0\\ \hline
		Gaussian gradient magnitude &  $\sigma$ & 0.5, 1.0, 1.5, 2.5, 3.5, 5.0\\ \hline
		Difference of Gaussians & $(\sigma_1,\sigma_2)$ & (1.0, 0.75), (1.5, 1.0), (2.5, 1.5)\\
		&&(3.5, 2.5), (5.0, 3.5)\\ \hline
		Hessian, Hessian eigenvalues & $\sigma$& 0.5, 0.75, 1.0 \\ \hline
		Structure tensor eigenvalues & $\sigma$ & 0.5, 0.75, 1.0\\ \hline
	\end{tabular}
	\caption{Set of image transforms used as training data for a random forest to classify images with cracks of width 3.}
	\label{rf:features}
\end{table}

\subsubsection*{3d U-Net (NN)}
To reduce the memory required for training the U-Net, the $256^3$ training images are split into patches of size $64^3$. The network is then trained on this smaller image size. Later, larger images can be processed by the network by splitting them into $64^3$ patches and merging the outputs after the classification. To prevent edge effects, the patches are chosen to overlap by 14 voxels.

Training a neural network for image segmentation requires the choice of a suitable loss function. In our application, we face the problem of class imbalance: The crack class (label 1) has much less voxels than the background class (label 0). In this case, it is harder to predict the minority class (the crack). To deal with this problem we use a weighted loss function. It is derived by weighting the ordinary binary cross entropy loss which is computed voxelwise between network output and target. The voxel losses are weighted before summing over the batch. 

The weights $w$ are calculated as follows:
Let $p_0$ and $p_1$ be the percentage of voxels in class 0 (background) and 1 (crack), respectively. The weight for the losses of crack voxels is then chosen as $w = p_0/p_1.$ The losses for background voxels are weighted by $w= 1$.
In this way, the two classes are treated as if they had the same number of voxels. 

To increase the size of the training set and the variety of crack images, a rather heavy data augmentation sequence is applied to each image. It includes geometric transformations like rotating and flipping the image as well as grayvalue shifts and distortions. This makes the network more robust. The sequence is applied to the training data twice, tripling its size. Thus, the network is trained on $768$ images of size $64^3$. Before training, all images are normalized.

It remains to specify the hyper-parameters of the network, namely batch size, number of epochs, and learning rate. Limited also by memory restrictions, we use a batch size of $2$ and $20$ epochs. 

Furthermore, we use an initial learning rate of $0.001$ with a decay of $0.5$ which is applied after every fifth epoch.

The only proper parameter of the method is the threshold $t_6\in[0,1]$ which is applied to the output image containing the class probabilities. We fix the threshold to $t_6=0.5$ for all crack widths.

\section{Comparison of the methods}\label{sec:comparison}

\subsection{Results}
All methods are applied to the validation data set by using the parameters derived in Section \ref{sec:parameters}. The values are listed in Appendix \ref{app:Para}. The results are evaluated by computing their precision, recall, and F1-score with a tolerance of $0$ and $1$. 

\subsubsection*{Optimizing precision:}

When maximizing precision, the goal is to minimize false positives, i.e. to extract voxels that have a very high probability of being contained in the crack. 
The cost of pursuing this objective is that some crack voxels may be missed. Maximizing precision minimizes noise in the results and, in some cases, extracts cracks that are too thin compared to the ground truth. 

\begin{figure}[h]
	\centering
	\includegraphics[width = .94\textwidth]{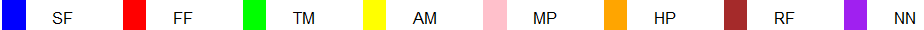}
	\includegraphics[width = .98\textwidth]{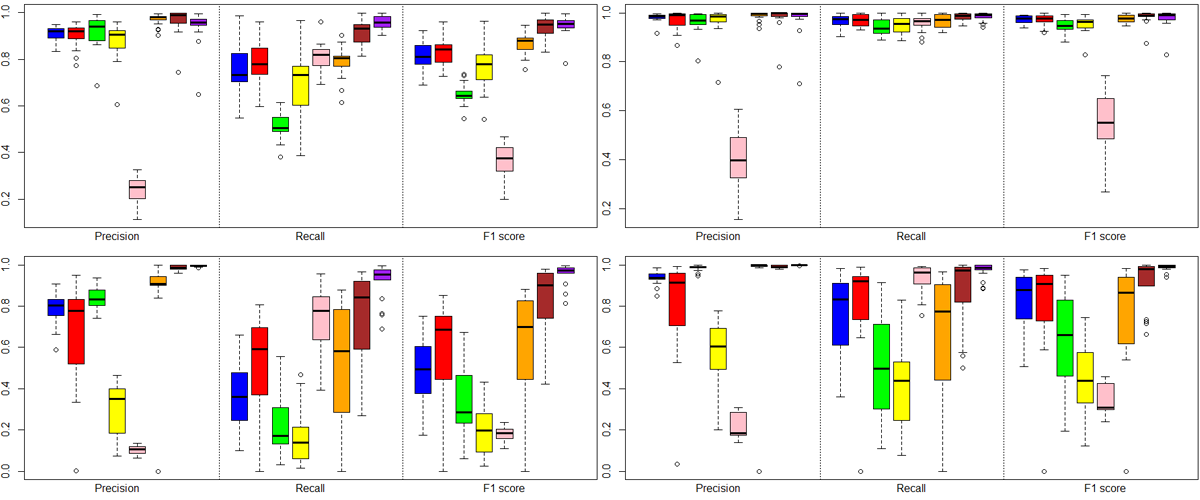}
	\caption{Results for precision optimization for tolerance $0$ (left) and $1$ (right) for crack width $3$ (top) and crack width $1$ (bottom).}
	\label{fig:precision-resw1w3}
\end{figure}

Figure~\ref{fig:precision-resw1w3} shows box plots of precision, recall, and F1-score for crack widths $3$ and $1$ and tolerances $0$ and $1$. The results for crack width $5$ are omitted because they are similar to those of crack width $3$. 
Selected 2d slices of the precision-optimized outputs are shown in Figure~\ref{fig:2d-visualized-all}. 3d volume renderings can be found in Figure~\ref{optPrec_9x9_w316}.

For images with crack width 3, the top three methods are RF, NN, and HP with median precision values above 0.95 without tolerance. Further analyzing these methods, we see that RF has the best precision values while NN has the best recall values which results in a similar F1-score. 

Among the classical methods, the HP algorithm yields the best results. Its precision values are close to those of RF (and even better than NN), but the recall (and thus F1-value) is significantly worse compared to the other two methods. Robustness of the methods can be deduced from their comparably small standard deviation.

Increasing the tolerance to $1$ voxel improves the results drastically. In this case, median recall values for all methods lie above 0.93. The precision and F1-values also increase. Except for MP, very high scores are obtained. The classical methods benefit more from increasing the tolerance level than the ML methods which produce good results already for $tol=0$. This also suggests that incorporating further post-processing or refinement into the classical methods could improve the results. MP achieves good recall values while its precision is by far the worst. In total, all other methods yield very good results for crack width $3$ when allowing a tolerance.

For crack width $1$, almost all methods perform worse than for crack width $3$. Again, the best results without tolerance are provided by NN, RF, and HP with median precision values above 0.90. Using a tolerance improves the results only  marginally. Regarding precision and recall, NN is slightly better than RF while HP performs worse than those two. For tolerance 1, we further note that TM can compete with respect to precision and MP with respect to recall. In general, standard deviations increase compared to the results of crack width 3, especially those of the recall values.

When evaluating the results visually, we notice that some methods still produce a large amount of false positives in the background. This explains the comparatively bad performance of MP
Also, the erroneous detection of air pores may deteriorate the results, for example in FF and AM. Most methods perform better on images with straight crack shapes. In TM and AM, this can be explained by the planar template/structuring element which is supposed to approximate the crack: If the crack structure is rather curved, we expect a worse approximation than for straight cracks.

\subsubsection*{Optimizing recall:}

\begin{figure}[h]
	\centering
	\includegraphics[width = .94\textwidth]{figs/labels2.png}
	\includegraphics[width = .98\textwidth]{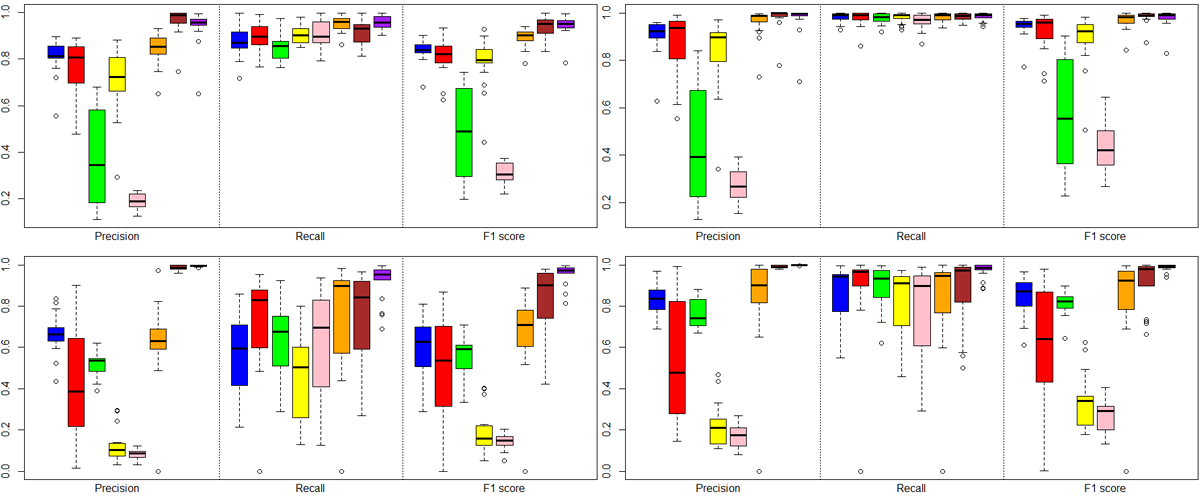}
	\caption{Results for recall optimization for tolerance $0$ (left) and $1$ (right) for crack width $3$ (top) and crack width $1$ (bottom).}
	\label{fig:recall-resw1w3}
\end{figure}

Here, the goal is to minimize false negatives, i.e. the focus is on covering the full crack structure voxelwise. For this objective, background voxels tend to be classified as cracks, especially in the proximity of the crack. The results for crack widths $3$ and $1$ are given in Figure~\ref{fig:recall-resw1w3}. Again, the results for crack width $5$ are omitted. For visual analysis, selected 2d slices are shown in Figure~\ref{fig:2d-visualized-all} and 3d volume renderings in Figure~\ref{optPrec_9x9_w316}.

For crack width $3$ and $tol=1$ the best performing methods are RF, HP, and NN with median recall values above 0.95.
Among these methods, RF has the highest precision. HP has the lowest precision but is still exceeding that of the other classical methods. NN is the most balanced method as mean precision and recall values are on the same level. It also yields the highest F1-score. Using $tol=1$ strongly increases the recall values while the precision values increase marginally, only.

With the exception of NN, all methods perform worse for crack width $1$. Again, the best methods are HP, RF, and NN both w.r.t. median recall (above 0.71) and median precision values (above 0.62). NN outperforms the other methods as it yields median recall and precision values above 0.91. 

With $tol=1$, the goodness of the classical methods improves significantly while the ML methods improve only slightly. Still, most classical methods are not competitive due to their poor precision values or a comparatively large standard deviation of the results. 

Air pores may influence the results negatively as cracks that go through pores are not detected at all. Furthermore, voxels on the edge of air pores may be falsely detected. 

\subsection{Robustness, limitations, and outliers}
All methods are very robust for crack widths $3$ and $5$. This can be concluded from the small standard deviations in the recall or precision values (depending on the objective) for $tol=1$. For crack width $1$, the situation is different. In the class of ML methods, RF performs worse compared to the other thickness levels but is still able to segment most of the crack voxels. Only NN is able to maintain its performance and even improves slightly.
The classical methods except HP completely fail on this task. A possible explanation is that those methods are not able to distinguish between thin crack structures and noise or texture originating from the concrete structure or the CT imaging. This problem is illustrated in Figure~\ref{fig:2d-visualized-all}. 

\begin{figure}[ht]
	\centering
	\includegraphics[width = 0.45\textwidth]{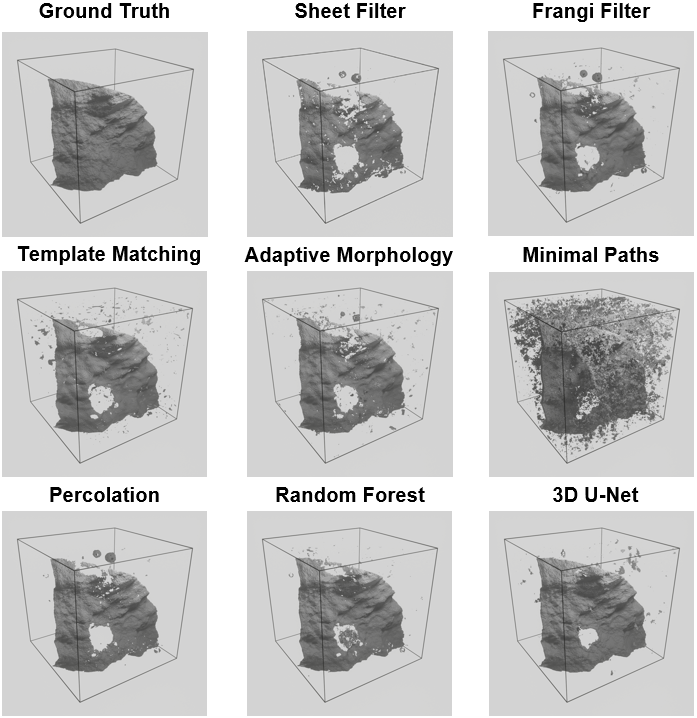}
	\qquad
	\includegraphics[width = 0.45\textwidth]{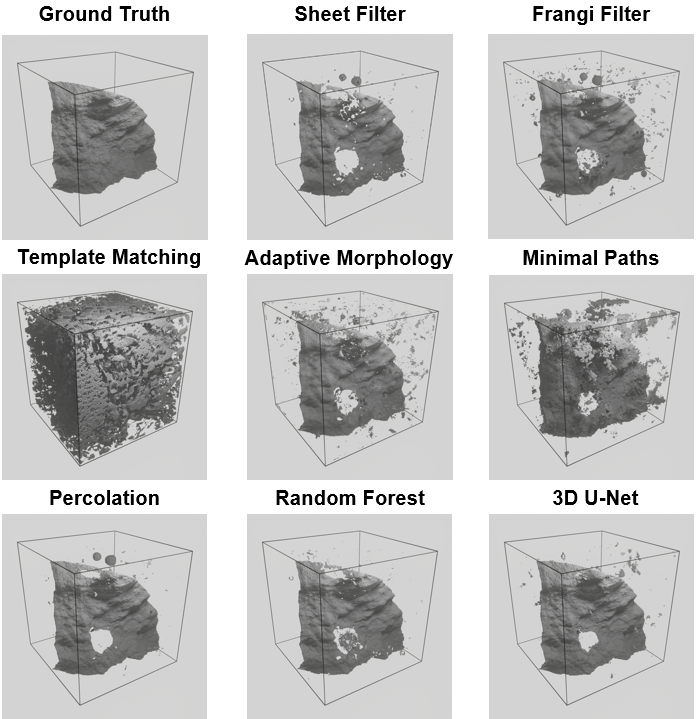}
	\caption{Renderings of ground truth and outputs of crack width 3. Left: parameters optimized with respect to precision, Right: parameters optimized with respect to recall.}
	\label{optPrec_9x9_w316}
\end{figure}

In cases where cracks propagate through air pores, most methods were not able to deduce the crack shape inside the pore but have correctly differentiated between crack and pore, see Figure~\ref{optPrec_9x9_w316}.

A further problem arises in case of low contrast between crack and background. 
Here, most methods fail to capture the full crack structure. Especially the methods based on the grayvalue distribution of the image perform worse than the ones based on geometric properties. In the example in the left of Figure~\ref{fig:2d-visualized-all}, only TM, MP, and NN detect the crack completely. The other methods miss parts of the crack in regions with lower contrast. 

Structures that are generally challenging for segmentation are curvy cracks and pore edges. 
For instance, TM and AM are based on shape matching. In curvy cracks, the assumptions on the crack shape hold only weakly. Hence, the methods are struggling to preserve connectivity. Pore edges are sometimes confused with cracks occasionally causing noisy segmentations. This is not a surprise since air pores have a similar grayvalue distribution as cracks. As a consequence, they may be considered to be part of a crack structure, for example by the edge detection filters.

\afterpage{
	\thispagestyle{empty}
	\begin{figure}[h]
		\centering
		\includegraphics[width = \textwidth]{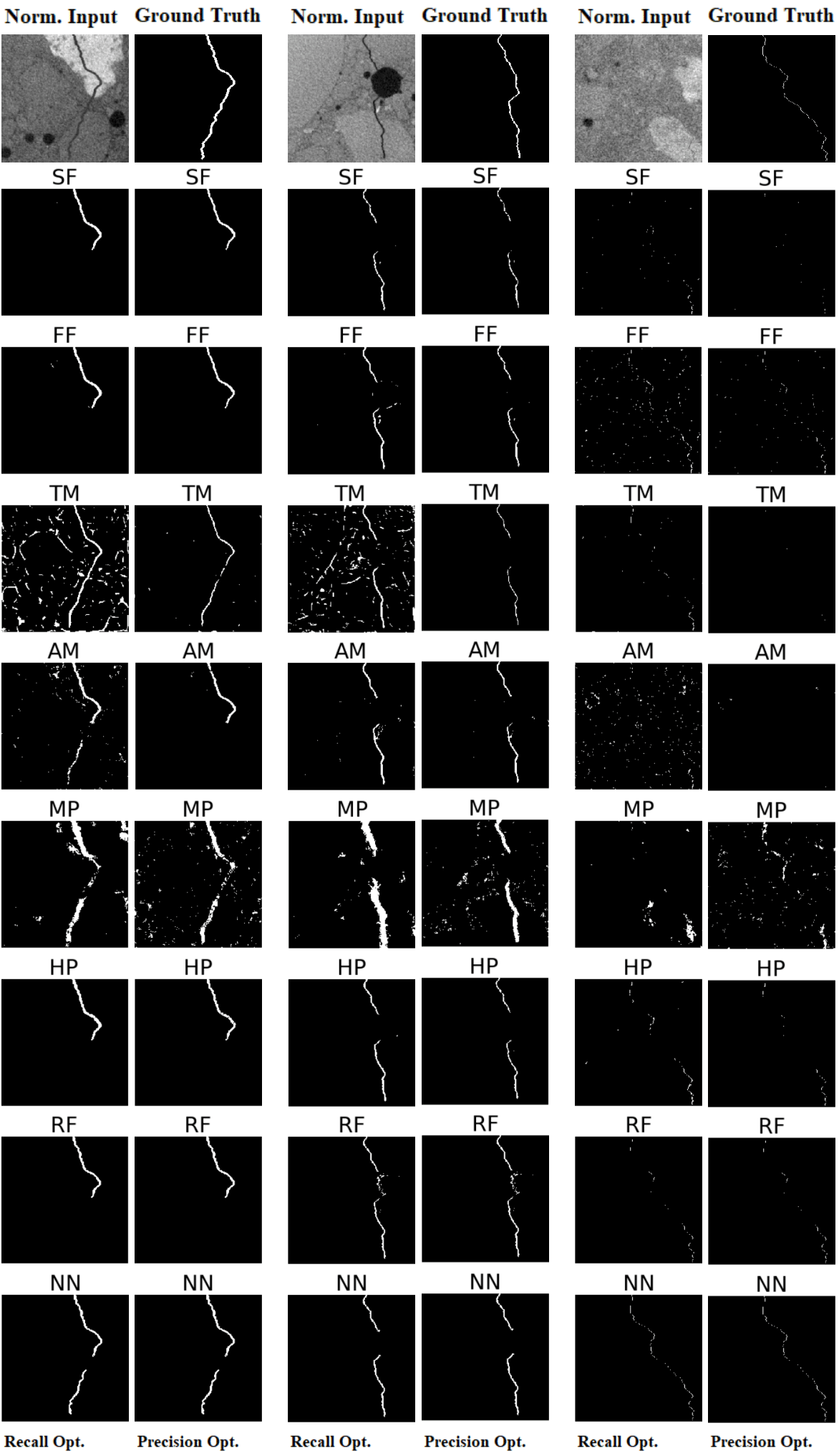}
		\caption{Selected 2d slices of outputs of the methods.}
		\label{fig:2d-visualized-all}
	\end{figure}
	\clearpage
}

\subsection{Run-time}
The run-times of the methods depend on several factors including image size, image structure, parameter choice, and size of training data. An extensive run-time comparison is beyond the scope of this paper. 
We only report rough estimates of the run-times on semi-synthetic images of size $256^3$. 

SF and FF are the fastest with run-times in the range of 20 seconds. However, we expect a strong increase when changing to a multi-scale approach. HP's run-time mainly depends on the size of the percolation window $W$. For the values considered here, HP takes 1-2 minutes. The run-time of MP increases with the path length. It ranges between 30 seconds and 4 minutes for a path length $\ell\in [12,48]$. TM is the slowest, taking  1 hour up to 8 hours depending on the discretization parameter. AM is significantly faster, taking around 2 minutes. As for TM, run-time is affected most by the discretization parameter, but also by the half-angle parameter. 

The learning methods require training before predicting. Training may take several days depending on the amount of training data. We sped up RF by parallelizing the computation of the decision trees resulting in  approximately 4 minutes for RF prediction for the parameters and training data chosen here. 
For NN using a patch sampling with an overlap of 14 voxels, the run-time for prediction is around 3 minutes.

\section{Conclusion and outlook}
\label{ch:conclusion}

We examine several methods for segmenting cracks in 3d images of concrete obtained by CT. The eight methods include approaches successfully applied on 2d crack images or to solve 3d segmentation tasks in other fields of research. These methods are adjusted or generalized to the specific problem of crack segmentation in 3d. 

We mainly distinguish between classical methods (sheet filter, Frangi filter, template matching, adaptive morphology, Hessian-based percolation, minimal paths algorithm) and learning methods (random forest, 3d U-Net). 

The methods are evaluated on semi-synthetic 3d crack images with varying crack widths. We conclude that the top three methods with respect to several performance measures studied in this paper are NN, RF, and HP. The mean F1-values are summarized in Table \ref{tab:mean-f1}. The results differ strongest for crack width $1$ where NN clearly performs best. The results of the top three methods are more similar for crack widths $3$ and $5$. The HP algorithm performs best among the classical methods but is still outperformed by both learning methods.

\begin{table}[ht]
	\centering
	\begin{tabular}{|c|c|c|c|c|}
		\hline
		Mean F1-values& crack width $1$ & crack width $3$ & crack width $5$\\
		\hline
		NN & 0.9509 (0.0387) & 0.9451 ( 0.0469) & 0.9334 (0.0539)\\
		RF & 0.7991  (0.1565)  & 0.9367  (0.0458) & 0.9193 (0.1353) \\
		HP (opt. precision) & 0.6228  (0.2433)  & 0.8675 (0.0458)  & 0.8615 (0.1257) \\
		HP (opt. recall) & 0.6652 (0.1997) & 0.8925 (0.0412)  & 0.8276 (0.1315) \\
		\hline
	\end{tabular}
	\caption{Mean F1-values over $17$ images per crack width. Standard deviation in brackets.}
	\label{tab:mean-f1}
\end{table}

The trade-off between achieving high precision and high recall is greater for classical methods than for learning methods. Hence, the parameters of these methods should be chosen depending on the particular goal of the segmentation. For the learning methods, this is not necessary. 

The RF is trained on individual voxels of the input image and selected image transforms. As a consequence, the performance of RF depends strongly on the choice of training data, and RF will only perform well on images with similar properties.
In terms of robustness against image diversity, NN seems to be the better choice. Data augmentation which increases the network's robustness can easily be incorporated in the training process. 

In summary, if suitable training data can be generated, the learning methods are clearly superior to the classical methods. Comparing random forest and 3d U-Net, we find that the 3d U-Net yields slightly better results, in particular in terms of robustness.

The use of classical methods is justified when semi-synthetic training data are not available. For images of new concrete types whose properties are not known a-priori, the effort of training a network may be too high. In this case, we recommend using the Hessian-based percolation algorithm.

The main goal of our future research is to find methods or procedures that enable crack segmentation for large CT scans 
($\sim 2\,000^3-2\,000^2\times10\,000$ pixels) of 
concrete with real cracks automatically or semi-automatically. Here, 
challenges arise from many sources, e.g., the complex structure and topology of cracks and the variety of concrete types. Additionally, run-time is more critical in these cases. One strategy to reduce the runtime could be to use a two stage approach. First, a computationally cheaper, rough method can be used to detect candidate regions that possibly contain cracks. Then, one of the methods presented here can be applied to those regions to obtain a precise crack segmentation.

Another issue in the real concrete images is that crack widths are not fixed and that cracks may transition from one scale to another while propagating through the concrete. Therefore, our methods should be generalized to account for multi-scale cracks. \newline

Funding: This work was supported by the German Federal Ministry of Education and Research (BMBF) [grant number 05M2020 (DAnoBi)].

\bibliographystyle{unsrt}
\bibliography{Methods-for-segmenting-cracks-in-3d-images-of-concrete-A-comparison-based-on-semi-synthetic-images.bib}

\bigskip
{\small
	\textbf{Tin Barisin} is a PhD student at the Fraunhofer ITWM (Image Processing Department) and the University of Kaiserslautern (Statistics Group), Germany. He holds bachelor's (Mathematics) and master's (Mathematical Statistics) degree from University of Zagreb, Croatia. His area of interest is mathematical image processing and mathematical morphology.
	\\
	\textbf{Christian Jung} is a PhD student at the University of Kaiserslautern, Statistics Group, Germany. Prior to that, he completed his bachelor's degree in Mathematics and his master's degree in Business Mathematics. His research interests include 3d image analysis and stochastic modelling of microstructures.
	\\
	\textbf{Franziska Müsebeck}  
	is a research assistant at the University of Kaiserslautern, Statistics Group. She completed her master's degree in Mathematics and minor in Computer Science at the University of Kaiserslautern in 2020. Her area of interest is image analysis and statistical learning. 
	\\
	\textbf{Prof. Dr. Claudia Redenbach} has held a professorship for statistics in the Mathematics Department of the University of Kaiserslautern since 2017. The focus of her research is on quantitative image analysis, spatial statistics, and stochastic microstructure modelling.
	\\
	\textbf{Dr. Katja Schladitz} is a mathematician with a PhD in stochastic geometry. She has worked in the image processing department at the Fraunhofer Institute of Industrial Mathematics ITWM in Kaiserslautern, Germany, since 1999. Her research and project work focuses on 3d image analysis and microstructure modelling.
}

\begin{appendices}
	
	\section{Random forest features}\label{app:RF}
	
	Let $I:\mathbb{R}^d \rightarrow \mathbb{R}$ be a d-dimensional image, $G:\mathbb{R}^d\times \mathbb{R}^+\rightarrow\mathbb{R}$ be a Gaussian kernel and, $\sigma,\sigma_1,\sigma_2\in\mathbb{R}_{>0}$. For every point $p$ in the domain of $I$, the Gaussian filter $F_{\sigma}$ is obtained by the convolution
	$F_{\sigma}\left(p\right) = \left(I * G\left(\cdot,\sigma\right)\right)\left(p\right),$ the difference of Gaussians is the pointwise difference of Gaussian filters, $F_{\sigma_1}(p)-F_{\sigma_2}(p)$,
	the Laplacian of Gaussian and the the Gaussian gradient magnitude are 
	$$I(p)* \left[ \sum_{i=1}^{d} \frac{\partial^2}{\partial x_i^2}G(p,\sigma) \right] \text{ and }I(p)* \sqrt{\sum_{i=1}^{d} \left( \frac{\partial}{\partial x_i}G(p,\sigma)\right)^2 },$$
	respectively, and for $d=3$ the structure tensor $T_{struct}$ of $I$ is given as
	\begin{equation*}
	T_{struct} (p,\sigma) = \left(a_{i,j}\right)^3_{i,j=1}, \text{\hspace{0.7cm}} a_{i,j} = (\frac{\partial}{\partial x_i} I (p,\sigma))\cdot (\frac{\partial}{\partial x_j} I (p,\sigma)).
	\end{equation*}
	
	\section{Parameter choices}\label{app:Para}
	Note that case $\sigma = 0$ in the Hessian matrix refers to the standard image gradient. 
	\begin{table}[h]
		\centering
		\resizebox{0.6\textwidth}{!}{%
			\begin{tabular}[H]{p{1cm}p{1cm}|p{1cm}|p{1cm}|p{1cm}|p{1cm}|p{1cm}|p{1cm}|} 
				\cline{3-8}
				\multicolumn{2}{c}{} & \multicolumn{3}{|c|}{Optimize precision} & \multicolumn{3}{c|}{Optimize recall}\\ 
				\cline{1-8}
				\multicolumn{2}{|c}{Width} & \multicolumn{1}{|c|}{5} & \multicolumn{1}{c|}{3} & \multicolumn{1}{c|}{1} & \multicolumn{1}{c|}{5} & \multicolumn{1}{c|}{3} & \multicolumn{1}{c|}{1}\\
				\cline{1-8}
				\multicolumn{1}{|l}{SF} & $\sigma$ 
				& 2.5 & 1.5 & 0.5 & 2.5 & 1.5 & 0.5\\
				\multicolumn{1}{|l}{} & $\rho$
				& 1 & 1 & 1 & 0.05 & 0.3 & 1\\
				\multicolumn{1}{|l}{} & $\delta$
				& 1 & 1.5 & 2.5 & 0.5 & 1 & 1.5\\
				\multicolumn{1}{|l}{} & $t_1$
				& 0.8 & 0.8 & 0.8 & 0.75 & 0.8 & 0.8\\
				\cline{1-8}
				\multicolumn{1}{|l}{FF} & $\sigma$ 
				& 2.5 & 1.5 & 0.5 & 2.5 & 1.5 & 0.5\\
				\multicolumn{1}{|l}{} & $\alpha$
				& 0.2 & 0.3 & 0.3 & 1 & 0.5 & 0.6\\
				\multicolumn{1}{|l}{} & $\beta$
				& 0.3 & 0.3 & 0.5 & 0.5 & 0.5 & 0.6\\
				\multicolumn{1}{|l}{} & $t_2$
				& 22 & 24 & 28 & 20 & 18 & 23\\
				\cline{1-8}
				\multicolumn{1}{|l}{TM} & $b$
				& 5 & 3 & 2 & 5 & 3 & 2\\
				\multicolumn{1}{|l}{} & $c$
				& 5 & 3 & 1 & 5 & 3 & 1\\
				\multicolumn{1}{|l}{} & $n$ 
				& 15 & 15 & 15 & 15 & 15 & 15\\
				\multicolumn{1}{|l}{} & $N$
				& 5 & 3 & 3 & 5 & 5 & 3\\
				\multicolumn{1}{|l}{} & $t_4$
				& 0.55 & 0.65 & 0.6 & 0.4 & 0.4 & 0.45\\
				\cline{1-8}
				\multicolumn{1}{|l}{AM} & $\sigma$ 
				& 1 & 1 & 0 & 1 & 1 & 0\\
				\multicolumn{1}{|l}{} & $\delta_{max}$
				& 0.5 & 0.5 & 0.5 & 0.5 & 0.5 & 0.5\\
				\multicolumn{1}{|l}{} & $n$
				& 20 & 20 & 20 & 20 & 20 & 20\\
				\multicolumn{1}{|l}{} & $N$
				& 3 & 3 & 3 & 3 & 3 & 3\\
				\multicolumn{1}{|l}{$k$} & (in $t_5$)
				& 4 & 4.5 & 4 & 2 & 2.5 & 2.5\\
				\cline{1-8}
				\multicolumn{1}{|l}{MP} & $\ell$ 
				& 12 & 12 & 12 & 48 & 48 & 48\\
				\multicolumn{1}{|l}{} & $t_3$
				& 0.0001 & 0.0001  & 0.0001 & 0.0001 & 0.0001 & 0.0001\\
				\cline{1-8}
				\multicolumn{1}{|l}{HP} & $\varepsilon$ 
				& -0.5 & -0.5 & -0.5 & -0.5 & -0.5 & -0.1\\
				\multicolumn{1}{|l}{} & $\tau$
				& 5 & 4 & 3 & 1 & 3 & 3\\
				\multicolumn{1}{|l}{} & $f$
				& 0.6 & 0.6 & 0.6 & 0.87 & 0.8 & 0.6\\
				\multicolumn{1}{|l}{} & $W$
				& 5 & 3 & 4 & 4 & 4 & 4\\
				\cline{1-8}
				\multicolumn{1}{|l}{RF} & $d_{dt}$ 
				& 50 & 50 & 50 & 50 & 50 & 50\\
				\multicolumn{1}{|l}{} & $n_{dt}$
				& 100 & 100 & 100 & 100 & 100 & 100\\
				\cline{1-8}
				\multicolumn{1}{|l}{NN} & $t_6$ 
				& 0.5 & 0.5 & 0.5 & 0.5 & 0.5 & 0.5\\
				\cline{1-8}
		\end{tabular}}
		\caption{Parameter choices for the methods evaluated on the test data set.}
	\end{table}
	
\end{appendices}

\end{document}